# On Interpretability of Artificial Neural Networks: A Survey

Feng-Lei Fan, *Student Member, IEEE*, Jinjun Xiong, *Senior Member, IEEE*, Mengzhou Li, *Student Member, IEEE*, and Ge Wang, *Fellow, IEEE*

*Abstract*— Deep learning as represented by the artificial deep neural networks (DNNs) has achieved great success recently in many important areas that deal with text, images, videos, graphs, and so on. However, the black-box nature of DNNs has become one of the primary obstacles for their wide adoption in mission-critical applications such as medical diagnosis and therapy. Because of the huge potentials of deep learning, increasing the interpretability of deep neural networks has recently attracted much research attention. In this paper, we propose a simple but comprehensive taxonomy for interpretability, systematically review recent studies in improving interpretability of neural networks, describe applications of interpretability in medicine, and discuss possible future research directions of interpretability, such as in relation to fuzzy logic and brain science.

## I. INTRODUCTION

Deep learning [71] has become the mainstream approach in many important domains targeting common objects such as text [40], images [182], videos [132], and graphs [88]. However, deep learning works as a black box model in the sense that, although deep learning performs quite well in practice, it is difficult to explain its underlying mechanism and behaviors. Questions are often asked such as how deep learning makes such a prediction, why some features are favored over others by a model, and what changes are needed to improve model performance, etc. Unfortunately, only modest success has been made to answer these questions.

Interpretability of deep neural networks is essential to many fields, and to healthcare [67], [68], [174] in particular for the following reasons. First, model robustness is a vital issue in medical applications. Recent studies suggest that model interpretability and robustness are closely connected [131]. On the one hand, the improvements in model robustness prompt model interpretability. For example, a deep model trained via adversarial training, a training method that augments training data with adversarial examples, shows better interpretability (with more accurate saliency maps) than the same model trained without adversarial examples [131]. On the other hand, when we understand a model deeply, we can thoroughly examine its weaknesses because the interpretability can help identify potential vulnerabilities of a complicated model, thereby improving its accuracy and reliability. Also, interpretability plays an important role in ethic use of deep learning techniques [57]. To build patients' trust in deep learning, interpretability is needed to hold a deep learning system accountable [57]. If a model builder can explain why a model makes a particular decision under certain conditions, users would know whether such a model contributes to an adverse event or not. It is then possible to establish standards and protocols to use the deep learning system optimally.

However, the lack of interpretability has become a main barrier of deep learning in its wide acceptance in mission-critical applications. For example, regulations were proposed by European Union in 2016 that individuals affected by algorithms have the right to obtain an explanation [61]. Despite great research efforts made on interpretability of deep learning and availability of several reviews on this topic, we believe that an up-to-date review is still needed, especially considering the rapid development of this area. The review of Q. Zhang and S. C. Zhu [202] is mainly on the visual interpretability. The representative publications from their review fall under the *feature analysis*, *saliency*, and *proxy* taxonomy in our review. The review of S. Chakraborty *et al.* [28] took opinions of [112] on levels of interpretability, and accordingly structured their review to provide in-depth perspectives but with limited scope. For example, only 49 references are cited there. The review of M. Du *et al.* [43] has a similar weakness, only covering 40 papers which are divided into post-hoc and ad-hoc explanations, as well as global and local interpretations. Their taxonomy is coarse-grained and neglects a number of important publications, such as publications on *explaining-by-text*, *explaining-by-case*, etc. In contrast, our review is much detailed and comprehensive, with the latest results included. While publications in L. H. Gilpin *et al.* [58] are classified into understanding the workflow of a neural network, understanding the representation of a neural network, and explanation producing, we cover all these aspects and also discuss the studies on how to protype an interpretable neural network. Reviews by R. Guidotti *et al.* [65] and A. Adadi and M. Berrada [2] cover existing black-box machine learning models instead of focusing on neural networks. As a result, several hallmark papers on explaining neural networks are missing in their survey, such as the interpretation from the perspective of mathematics and physics.

A. B. Arrieta *et al.* [10] provides an extensive review on explainable AI (XAI), where concepts and taxonomies are clarified, and challenges are identified. While that review covers interpretability of AI/ML in general, our review is specific to deep neural networks and offers unique perspectives and insights. Specifically, our review is novel in the following senses: 1) We treat post-hoc and ad-hoc interpretability separately，because the former explains the existing models, while the latter constructs interpretable ones; 2) we include widely-studied generative models, advanced

This work was supported in part by the Rensselaer-IBM AI Research Collaboration Program (http://airc.rpi.edu), part of the IBM AI Horizons Network (http://ibm.biz/AIHorizons)

Feng-Lei Fan, Mengzhou Li, and Ge Wang* (wangg6@rpi.edu) are with Department of Biomedical Engineering, Rensselaer Polytechnic Institute, Troy, NY, USA. Asterisk indicates the corresponding author.
Jinjun Xiong is with IBM Thomas J. Watson Research Center, Yorktown Heights, NY, 10598, USA. (email: jinjun@us.ibm.com )

mathematical/physical methods that summarize advances in deep learning theory, and the applications of interpretability in medicine; 3) important methods are illustrated with customized examples and publicly available codes through GitHub; and 4) interpretability research is a rapidly evolving field, and many research articles are published every year. Hence, our review should be a valuable and up-to-date addition to the literature.

Before we start our survey, let us first state three essential questions regarding interpretability: What does interpretability mean? Why is interpretability difficult? And how to build a good interpretation method? The first question has been well addressed in [112], and we include their statements here for completeness. The second question was partially touched in [112], [146], and we incorporate those comments and complement them with our own views. We provide our own perspectives on the third question.

*A. What Does Interpretability Mean?*

Although the word "interpretability" is frequently used, people do not reach a consensus on the exact meanings of interpretability, which partially accounts for why current interpretation methods are so diverse. For example, some researchers explore post-hoc explanations for models, while some focus on the interplay mechanism between machineries of a model. Generally speaking, interpretability refers to the extent of human's ability to understand and reason a model. Based on the categorization of [112], we summarize the implications of interpretability in different levels.

· *Simulatability* is considered as the understanding over the entire model. In a good sense, we can understand the mechanism of a model at the top level in a unified theoretical framework, one example is what was reported in [140]: a class of radial basis function (RBF) networks can be expressed by a solution to the interpolation problem with a regularization term, where a RBF network is an artificial neural network with RBFs as activation functions. In view of simulatability, the simpler the model is, the higher simulatability the model has. For example, a linear classifier or regressor is totally understandable. To enhance simulatability, we can change some facilities of models or use crafted regularization terms.

· *Decomposability* is to understand a model in terms of its components such as neurons, layers, blocks, and so on. Such a modularized analysis is quite popular in engineering fields. For instance, the inner working of a complicated system is factorized as a combination of functionalized modules. A myriad of engineering examples such as software development and optical system design have justified that a modularized analysis is effective. In machine learning, a decision tree is a kind of modularized methods, where each node has an explicit utility to judge if a discriminative condition is satisfied or not, each branch delivers an output of a judgement, and each leaf node represents the final decision after computing all attributes. Modularizing a neural network is advantageous to the optimization of the network design since we know the role of each and every component of the entire model.

· *Algorithmic Transparency* is to understand the training process and dynamics of a model. The landscape of the objective function of a neural network is highly non-convex. The fact that deep models do not have a unique solution hurts the model transparency. Nevertheless, it is intriguing that current stochastic gradient descent (SGD)-based learning algorithms still perform efficiently and effectively. If we can understand why learning algorithms work, deep learning research and applications will be accelerated.

*B. Why Is Interpretability Difficult?*

After we learn the meanings of interpretability, a question is what obstructs practitioners to obtain interpretability. This question was partially addressed in [146] in terms of *commercial barrier* and *data wildness*. Here, we complement their opinion with additional aspects on *human limitation* and *algorithmic complexity*. We believe that the hurdles to interpretable neural networks come from the following four aspects.

· *Human Limitation*: Expertise is often insufficient in many applications. Nowadays, deep learning has been extensively used in tackling intricate problems, which even professionals are unable to comprehend adequately. What's worse is that these problems are not uncommon. For example, in a recent study [46], we proposed to use an artificial neural network to predict pseudo-random events Specifically, we fed 100,000 binary sequential digits into the network to predict the $100,001^{th}$ digit in the sequence. In our prediction, the highly sophisticated hidden relationship was learned to beat a purely random guess with a 3σ precision. Furthermore, it was conjectured that high sensitivity and efficiency of neural networks may help discriminate the fundamental differences between pseudo-randomness and real quantum randomness. In this case, it is no wonder that interpretability for neural networks will be missing, because even most talented physicists know little about the essence of this problem, let alone fully understand predictions of the neural network.

· *Commercial Barrier:* In the commercial world, there are strong motives for corporations to hide their models. First and foremost, companies profit from black-box models. It is not a common practice that a company makes capital out of totally transparent models [146]. Second, model opacity helps protect hard work from being reverse engineered. An effective black box is ideal in the sense that customers being served can obtain satisfactory results while competitors are not able to steal their intellectual properties easily [146]. Third, prototyping an interpretable model may cost too much in terms of financial, computational, and other resources. Existing open-sourced superior models are accessible to easily construct a well-performed algorithm for a specific task. However, generating reliable and consistent understanding to the behavior of the resultant model demands much more endeavors.

· *Data Wildness:* On the one hand, although it is a big data era, high quality data are often not accessible in many domains. For example, in the project of predicting electricity grid failure [146], the data base involves text documents, accounting data about electricity dating back to 1890s, and data from new manhole inspections. Highly heterogenous and inconsistent data hamper not only the accuracy of deep learning models but also the construction of interpretability. On the other hand, real-world data have the character of high dimensionality, which suppresses reasoning. For example, given an MNIST image classification problem, the input image is of size $28 \times 28 = 784$. Hence the deep learning model tackling this problem has to learn an effective mapping of 784 variables to one of ten digits. If we consider the ImageNet dataset, the number of input variables goes up to $512 \times 512 \times 3 = 768432$.

· *Algorithmic Complexity*: Deep learning is a kind of large-scale, highly nonlinear algorithms. Convolution, pooling, nonlinear activation, shortcuts, and so on contribute to variability of neural networks. The number of trainable parameters of a deep model can be on the order of hundreds million or even more. Despite that nonlinearity may not necessarily result in opacity (for example, a decision tree model is not linear but interpretable), deep learning's series of nonlinear operations indeed prevent us from understanding its inner working. In addition, recursiveness is another source of difficulty. A typical example is the chaos behavior resultant from nonlinear recursiveness. It is well-known that even a simple recursive mathematical model can lead to intractable dynamics [107]. In [175], it was proved that there are chaotic behaviors such as bifurcations even in simple neural networks. In chaotic systems, tiny changes of initial inputs may lead to huge outcome differences, adding to the complexity of interpretation methods.

*C. How to Build a Good Interpretation Method?*

The third major issue is the criteria for assessing quality of a proposed interpretability method. Because existing evaluation methods are still premature, we propose five general and well-defined rules-of-thumb: *exactness*, *consistency*, *completeness*, *universality*, and *reward*. Our rules-of-thumb are fine-grained and focus on the characteristics of interpretation methods, compared to that described in [42]: application-grounded, human-grounded, and function-grounded.

· *Exactness:* Exactness means how accurate an interpretation method is. Is it just limited to a qualitive description or with a quantitative analysis? Generally, quantitative interpretation methods are more desirable than qualitative counterparts.

· *Consistency:* Consistency suggests that there is not any contradiction in an explanation. For multiple similar samples, a fair interpretation should produce consistent answers. In addition, an interpretation method should conform to the predictions of the authentic model. For example, the proxy-based methods are evaluated based on how closely they replicate the original golden model.

· *Completeness:* Mathematically, a neural network is to learn a mapping that best fits data. A good interpretation method should show effectiveness in support of the maximal number of data instances and data types.

· *Universality:* With the rapid development of deep learning, the deep learning armory has been substantially enriched. Such diverse deep learning models play important roles in a wide spectrum of applications. A driving question is whether we can develop a universal interpreter that deciphers as many models as possible so as to save labor and time. But this is technically challenging due to the high variability among models.

· *Reward:* What are gains from the improved understanding of neural networks? In addition to the trust from practitioners and users, fruits of interpretability can be insights into network design, training, etc. Due to its black-box nature, using neural networks is largely a trial-and-error process with sometimes contradictive intuitions. A thorough understanding of deep learning will be instrumental to the research and applications of neural networks.

Briefly, our contributions in this review are three-folds: 1) We propose a comprehensive taxonomy for interpretability of neural networks and describe key methods with our insights; 2) we systematically illustrate interpretability methods as educational aids, as shown in Figures 3, 5, 6, 7, 9, 10, 16, 17; and 3) we shed light on future directions of interpretability research in terms of the convergence of neural networks and rule systems, the synergy between neural networks and brain science, and interpretability in medicine.

## II. A SURVEY ON INTERPRETATION METHODS

In this section, we first present our taxonomy and then review interpretability results under each category of our taxonomy. We enter the search terms "Deep Learning Interpretability", "Neural Network Interpretability", "Explainable Neural Network", and "Explainable Deep Learning" into the Web of Science on Sep 22, 2020, with the time range from 2000 to 2019. The number of articles with respect to years is plotted in Figure 1, which clearly shows an exponential trend in this field. With the survey, our motive is to cover as many important papers as possible, and therefore we do not limit ourselves within Web of Science. We also search related articles using Google Scholar, PubMed, IEEE Xplore, and so on.

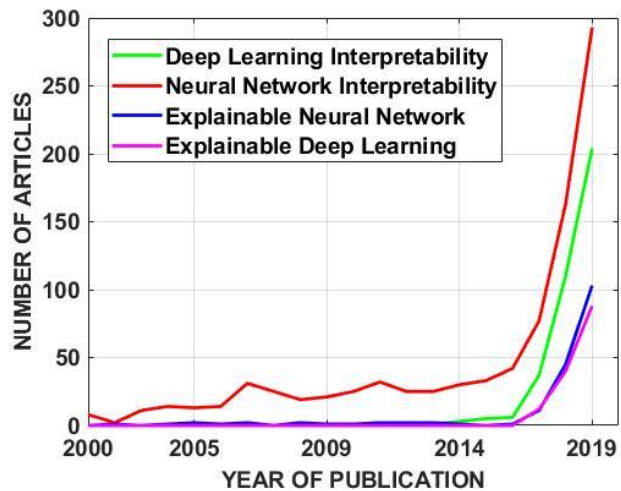

Figure 1. Exponential growth of the number of articles on interpretability.

*A. Taxonomy Definition*

As shown in Figure 2, our taxonomy is based on our surveyed papers and existing taxonomies. We first classify the surveyed papers into post-hoc interpretability analysis and ad-hoc interpretable modeling. Post-hoc interpretability analysis explains existing models and can be further classified into *feature analysis*, *model inspection*, *saliency*, *proxy*, *advanced mathematical/physical analysis*, *explaining-by-case*, and *explaining-by-text*, respectively. Ad-hoc interpretable modeling builds interpretable models and can be further categorized into *interpretable representation* and *model renovation*. In our proposed taxonomy, the class "*advanced mathematical/physical analysis*" is novel, but it is unfortunately missing in the previous reviews. We argue that this class is rather essential, because the incorporation of math/physics is critical in placing deep learning on a solid foundation for interpretability. In the following, we clarify the taxonomy definition and its illustration. We would like to underscore that one method may fall into different classes, depending on how one views it.

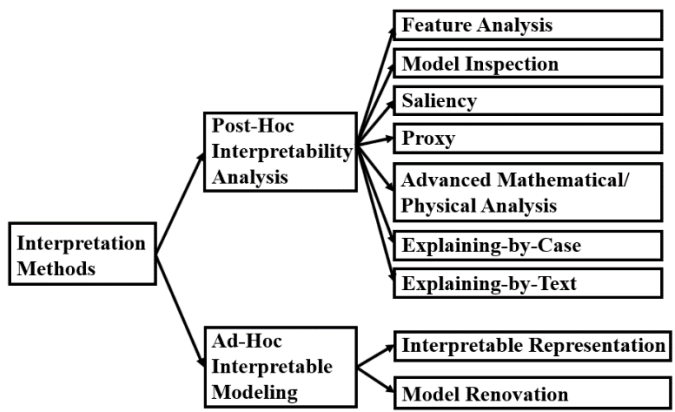

Figure 2. Taxonomy used for this interpretability review.

· *Post-hoc Interpretability Analysis*

Post-hoc interpretability is conducted after a model is well learned. A main advantage of post-hoc methods is that one does not need to compromise interpretability with the predictive performance since prediction and interpretation are two separate processes without mutual interference. However, a post-hoc interpretation is usually not completely faithful to the original model. If an interpretation is 100% accurate compared to the original model, it becomes the original model. Therefore, any interpretation method in this category is more or less inaccurate. What is worse is that we often do not know the nuance [146]. Such a nuance makes it hard for practitioners to have a full trust to an interpretation method, because the correctness of the interpretation method is not guaranteed.

*Feature analysis* techniques are centered in comparing, analyzing, and visualizing features of neurons and layers. Through feature analysis, sensitive features and ways to process them are identified such that the rationale of the model can be explained to some extent.

Feature analysis techniques can be applied to any neural networks and provide qualitative insights on what kinds of features are learned by a network. However, these techniques lack an in-depth, rigorous, and unified understanding, and therefore cannot be used to revise a model towards a higher interpretability.

*Model inspection* methods use external algorithms to delve into neural networks by systematically extracting important structural and parametric information on inner working mechanisms of neural networks.

Methods in this class are more technically accountable than those in *feature analysis* because analytical tools such as statistics are directly involved in the performance analysis. Therefore, the information gained by a model inspection method is more trustworthy and rewarding. In an exemplary study [184], finding important data routing paths is used as a way to understand the model. With such data routing paths, the model can be faithfully compressed to a compact one. In other words, interpretability improves the trustworthiness of model compression.

*Saliency* methods identify which attributes of input data are most relevant to a prediction or a latent representation of a model. In this category, human inspection is involved to decide if a saliency map is plausible. A saliency map is useful. For example, if a polar bear always appears in a picture coupled with snow or ice, the model may have misused the information of snow or ice to detect the polar bear rather than real features of polar bears for detection. With a saliency map, this issue can be found and hence avoided.

Saliency methods are popular in interpretability research, however, extensive random tests reported that some saliency methods can be model independent and data independent [3], i.e., saliency maps offered by some methods can be highly similar to results produced with edge detectors. This is problematic because it means that those saliency methods fail to find the true attributes of the input that account for the prediction of the model. Consequently, a model-relevant and data-relevant saliency method should be developed in these cases.

*Proxy* methods construct a simpler and more interpretable proxy that closely resembles a trained, large, complex, and black-box deep learning model. Proxy methods can be either local in a partial space or global in a whole solution space. The exemplary proxy models include decision trees, rule systems, and so on. The weakness of proxy methods is the extra cost needed to construct a proxy model.

*Advanced mathematical/physical analysis* methods put a neural network into a theoretical mathematics/physics framework, in which the mechanism of a neural network is understood with advanced mathematics/physics tools. This class covers theoretical advances of deep learning including non-convex optimization, representational power, and generalization ability.

A concern in this class is that, to establish a reasonable interpretation, unrealistic assumptions are sometimes made to facilitate a theoretical analysis, which may compromise the practical validity of the explanation.

*Explaining-by-case* methods are along the line of case-based reasoning [90]. People favor examples. One may not be engaged by boring statistic numbers of a product but could be amazed while listening to other users' experience of using such a product. This philosophy wins the heart of many practitioners and intrigues the case-based interpretation for deep learning. Explaining-by-case methods provide representative examples that capture the essence of a model.

Methods in this class are interesting and inspiring. However, this practice is more like a sanity check instead of a general interpretation because not much information regarding the inner working of a neural network is understood from selected query cases.

*Explaining-by-text* methods generate text descriptions in image-language joint tasks that are conducive to understanding the behavior of a model. This class can also include methods that generate symbols for explanation.

Methods in this class are particularly useful in image-language joint tasks such as generating a diagnostic report from an X-ray radiograph. However, explaining-by-text is not a general technique for any deep learning model because it can only work when a language module exists in a model.

· Ad-hoc Interpretable Modeling

Ad-hoc interpretable modeling eliminates the biases from the post-hoc interpretability analysis. Although it is generally believed that there is a trade-off between interpretability and model expressibility [123], it is still possible to find a model that is both powerful and interpretable. One notable example is the work reported in [30], where an interpretable two-layer additive risk model has won the first place in FICO Recognition Contest.

*Interpretable representation* methods employ regularization techniques to steer the optimization of a neural network towards a more interpretable representation. Properties such as decomposability, sparsity, and monotonicity can enhance interpretability. As a result, regularized features become a way to allow more interpretable models. Correspondingly, the loss function must contain a regularization term for the purpose of interpretability, which restricts the original model to perform its full learning task.

*Model renovation* methods seek interpretability by the means of designing and deploying more interpretable machineries into a network. Those machineries include a neuron with purposely designed activation function, an inserted layer with a special functionality, a modularized architecture, and so on. The future direction is to use more and more explainable components that can at the same time achieve similar state-of-the-art performance for diverse tasks.

B. Post-hoc Interpretability Analysis

· Feature Analysis

Inverting-based methods [41], [117], [164], [201] crack the representation of a neural network by inverting feature maps into a synthesized image. For example, A. Mahendran and A. Vedaldi [117] assumed that a representation of a neural network $\Omega_0$ for an input image $x_0$ was modeled as $\Omega_0 = \Omega(x_0)$, where $\Omega$ is the neural network mapping, usually not invertible. Then, the inverting problem was formulated as finding an image $x^*$ whose neural network representation best matches $\Omega_0$, i.e., $\arg\min_x \left\|\Omega(x) - \Omega_0\right\|^2 + \lambda R(x)$, where $R(x)$ is a regularization term representing prior knowledge about the input image. The goal is to reveal the lost information by comparing differences between the inverted image and the original one. A. Dosovitskiy *et al.* [41] directly trained a new network with features generated by the model of interest as the input and images as the label, to invert features of intermediate layers to images. It was found that contours and colors could still be reconstructed even from deeper layer features. M. D. Zeiler *et al.* [201] designed a deconvolution network consisting of unpooling, rectification, deconvolution operations, to pair with the original convolutional network so that features could be inverted without training. In the deconvolution network, an unpooling layer is realized by using locations of maxima; rectification is realized by setting negative values to zero; and deconvolution layers use transposed filters.

Activation maximization methods [45], [128], [129], [169] devote to synthesizing images that maximize the output of a neural network or neurons of interest. The resulting images are referred as "deep dreams" as these can be regarded as dream images of a neural network or a neuron.

In [16], [85], [108], [197], [211], it was pointed out that information about a deep model could be extracted from each neuron. J. Yosinski *et al.* [197] straightforwardly inspected the activation values of neurons in each layer with respect to different images or videos. They found that live activation values that change for different inputs are helpful to understand how a model work. Y. Li *et al.* [108] contrasted features generated by different initializations to investigate if a neural network learns a similar representation when randomly initialized. The receptive field (RF) is a spatial extent over which a neuron connects with an input volume [111]. To investigate the size and shape of RF of a given input for a neuron, B. Zhou *et al.* [211] presented a network dissection method that first selected $K$ images with high activation values for neurons of interest and then constructed 5,000 occluded images for each of $K$ images, and then fed them into a neural network to observe the changes in activation values for a given unit. A large discrepancy signals an important patch. Finally, the occluded images that have large discrepancy were re-centered and averaged to generate an RF. This network dissection method has been scaled to generative networks [17]. In addition, D. Bau *et al.* [16] scaled up a low-resolution activation map of a given layer to the same size as the input, thresholded the map into a binary activation map, and then computed the overlapping area between the binary activation map and the ground-truth binary segmentation map as an interpretability measure. A. Karpathy *et al.* [85] defined the gate in LSTM [73] to be either left or right saturated depending on its activation value being either less than 0.1 or more than 0.9. In this regard, neurons that are often right saturated are interesting because this means that these neurons can remember their values over a long period. Q. Zhang *et al.* [203] dissected feature relations in a network with the premise that the feature map of a filter in each layer can be activated by part patterns in the earlier layer. They mined part patterns layer by layer, discovered activation peaks of part patterns from the feature map of each layer, and constructed an explanatory graph to describe the relations of hierarchical features, with each node representing a part pattern and the edge between neighboring layers representing a co-activation relation.

· *Model Inspection*

The empirical influence function is to measure the dependence of an estimator on a sample [99]. P. W. Koh and P. Liang [89] applied the concept of the influence function to address the following question: Given a prediction for one sample, do other samples in the dataset have positive effects or negative effects on that prediction? This analysis could also help identify mis-annotated labels and outliers existing in the data. As Figure 3 shows, given a LeNet-5 like network, two harmful images for a given image are identified by the influence function.

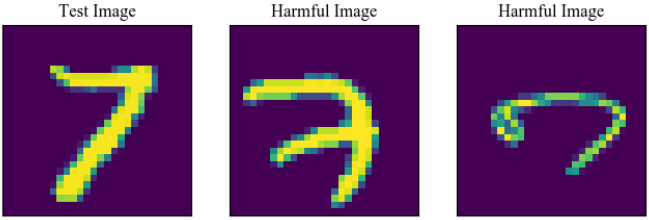

Figure 3. Based on the influence function, two harmful images that have the same label as the test image are identified.

A. Bansal *et al.* [12], H. Lakkaraju *et al.* [97], and Q. Zhang *et al.* [204] worked on the detection of failures or biases in a neural network. For example, A. Bansal *et al.* [12] developed a model-agnostic algorithm to identify which instances a neural network is likely to fail to provide any prediction for. In such a scenario, the model would instead give a warning like "Do not trust these predictions" as an alert. Specifically, they annotated all failed images with a collection of binary attributes and clustered these images in the attribute space. As a result, each cluster indicates a failure mode. To recognize those mislabeled instances with high predictive scores in the dataset efficiently, H. Lakkaraju *et al.* [97] introduced two basic speculations: The first is that mislabeling an instance with high confidence is due to the systematic biases instead of random perturbation, while the second is that each failed example is representative and informative enough. Then, they clustered the images into several groups and designed a multi-armed bandit search strategy by taking each group as a bandit that plans which group should be queried and sampled in each step. To discover representation biases, Q. Zhang *et al.* [204] utilized ground-truth relationships among attributes according to human's common knowledge (fire-hot vs ice-cold) to examine if a mined attribute relationship by a neural network well fits the ground truth.

Y. Wang *et al.* [184] demystified a network by identifying critical data routes. Specifically, a gate control binary vector $\lambda_k \in \{0,1\}^{n_k}$, where $n_k$ is the number of neurons in the $k^{th}$ layer, was multiplied to the output of the $k^{th}$ layer, and the problem of finding control gate values is formulated as searching $\lambda_1, \ldots, \lambda_K$:

$$\arg\min_{\lambda_1,\ldots,\lambda_K} d\big(f_\theta(x), f_\theta(x; \lambda_1, \ldots, \lambda_K)\big) + \gamma \sum_k ||\lambda_k||_1,$$

where $f_\theta$ is the mapping represented by a neural network parameterized by $\theta$, $f_\theta(x; \lambda_1, \ldots, \lambda_K)$ is the mapping when control gates $\lambda_1, \ldots, \lambda_K$ are enforced, $d(\cdot,\cdot)$ is a distance measure, $\gamma$ is a constant controlling the trade-off between the loss and regularization, and $||\cdot||_1$ is the $l_1$ norm such that $\lambda_k$ is sparse. The learned control gates could expose the important data processing paths of a model. B. Kim *et al.* [86] developed the concept activated vector (CAV) that can quantitively measure the sensitivity of the concept $C$ with respect to any layer of a model. First, a binary linear classifier $h$ was trained to distinguish between layer activations stimulated by two sets of samples: $\{f_l(x): x \in P_C\}$ and $\{f_l(x): x \notin P_C\}$, where $f_l(x)$ is the layer activation at the $l^{th}$ layer, and $P_C$ denotes data embodying the concept $C$. Then, the CAV was defined as the normal unit vector $v_C^l$ to a hyperplane of the linear classifier that separated samples with and without the defined concept. Finally, $v_C^l$ was used to calculate the sensitivity for a concept $C$ in the $l^{th}$ layer as the directional derivatives:

$$S_{C,k,l} = \lim_{\epsilon \to 0} \frac{h_{l,k}\big(f_l(x) + \epsilon v_C^l\big) - h_{l,k}(f_l(x))}{\epsilon} = \nabla h_{l,k}(f_l(x)) v_C^l,$$

where $h_{l,k}$ denotes the logits of the trained binary linear classifier for the output class $k$. J. You *et al.* [196] mapped a neural network into a relational graph, and then studied the relationship between the graph structures of neural networks and their predictive performance through massive experiments (transcribed a graph into a network and implemented the network on a dataset). They discovered that the predictive performance of a network was correlated with two graph measures: the clustering coefficient and the average path length.

· *Saliency*

There is a plethora of methods to obtain a saliency map. Partial dependence plot (PDP) and individual condition expectation (ICE) [53], [59], [74] are model-agnostic statistical tools to visualize the dependence between the responsible variables and the predictive variables. To compute the PDP, suppose there are $p$ input dimensions and let $S, C \subseteq \{1,2,..p\}$ be two complementary sets, where $S$ is the set one will fix, and $C$ is the set one will change. Then the PDP for $x_S$ is defined by $f_S = \int f(x_S, x_C) dx_C$, where $f$ is the model. Compared with PDP, the definition of ICE is straightforward. The ICE curve at $x_S$ is obtained by fixing $x_C$ and varying $x_S$. Figure 4 shows a simple example on how to compute PDP and ICE, respectively.

A simple approach is to study the change of prediction after removing one feature, also known as leave-one-out attribution [4], [83], [105], [143], [212]. For example, A. Kádár *et al.* [83] utilized this idea to define an omission score: $1 - cosine(h(S), h(S_{\backslash i}))$, where $cosine(\cdot,\cdot)$ is the cosine distance, $h$ is the representation for a sentence, $S$ is the full sentence, and $S_{\backslash i}$ is the sentence without the $i^{th}$ word, and analyzed the importance of each word. P. Adler *et al.* [4] proposed to measure an indirect influence for correlated inputs. For example, in a house loan decision system, race should not be a factor for decision-making. However, solely removing the race factor is not sufficient to rule out the effect of race because some remaining factors such as "zipcode" are highly concerned with race.

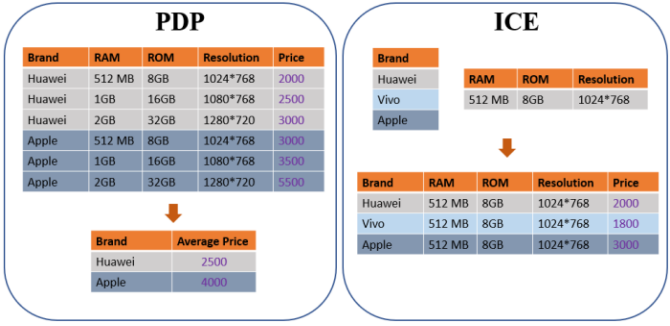

Figure 4. Toy examples illustrating the definitions of PDP and ICE, respectively. On the left, to measure the impact of the brand on the price with the PDP method, we fix the brand and compute the average of prices as other factors change, obtaining that the PDP of "Huawei" is 2500 and the PDP of "Apple" is 4000. On the right, ICE scores regarding brands "Huawei", "Vivo" and "Apple" are computed by varying brands and fixing other factors.

Furthermore, Shapley value from cooperative game theory was used in [6], [27], [39], [113], [115]. Mathematically, Shapley value of a set function $\hat{f}$ with respect to the feature $i$ is defined as

$$\text{Shapley}_i(\hat{f}) = \sum_{S \subseteq P \setminus \{i\}} \frac{(N-|S|-1)!|S|!}{N!} (\hat{f}(S \cup \{i\}) - \hat{f}(S)),$$

where $|\cdot|$ is the size of a set, $P$ is a total player set of $N$ players, and the set function $\hat{f}$ maps each subset $S \subseteq P$ to a real number. Furthermore, the definition of Shapley value can be twisted to the neural network function $f$ by replacing the features in the input that are not in $S$ with the zero value. Motivated by reducing the prohibitive computational cost incurred by combinatorial explosion, M. Ancona et al. [6] proposed a novel and polynomial-time approximation for Shapley values, which basically computed the expectation of a random coalition rather than enumerated each and every coalition. Figure 5 shows a simple example of how Shapley values can be computed for a fully connected layer network trained on California Housing dataset which includes eight attributes such as house age and room number as the inputs and the house price as the label.

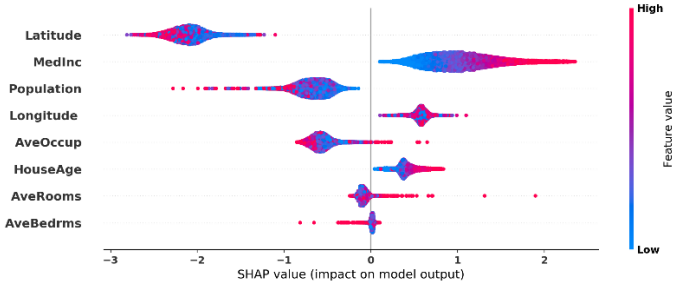

Figure 5. Positive Shapley value indicates a positive impact on the model output, and vice versa. Shapley value analysis shows that the model is biased because the house age has the positive Shapley value on the house price, which goes against with our real experience.

Instead of removing one or more features, researchers also resort to gradients. K. Simonyan et al. [157], D. Smilkov et al. [161], M. Sundararajan et al. [168] and S. Singla et al. [160] utilized the idea of gradients to probe the saliency of an input.

K. Simonyan et al. [157] calculated the first-order Taylor expansion of the class score with respect to image pixels, by which the first-order coefficients produce a saliency map for a class. D. Smilkov et al. [161] demonstrated that gradients as a saliency map show a correlation between attributes and labels, however, typically gradients are rather noisy. To remove noise, they proposed "SmoothGrad" that adds noise into the input image multiple times and averages the resultant gradient maps: $\widehat{M_c}(x) = \frac{1}{N} \sum_{n=1}^{N} M_c^{(n)}(x + N(0, \sigma^2))$, where $M_c^{(n)}$ is a gradient map for a class c, and $N(0, \sigma^2)$ is the Gaussian noise with $\sigma$ as the standard variance. Basically, $\widehat{M_c}(x)$ is a smoothened version of a salient map. M. Sundararajan et al. [168] set two fundamental requirements for saliency methods: (sensitivity) if only one feature is different between the input and the baseline, and the outputs of the input and the baseline are different, then this very feature should be credited by a non-zero attribution; (implementation invariance) the attributions for the same feature in two functionally equivalent networks should be identical. Noticing that earlier gradient-based saliency methods fail the above two requirements, they put forth integrated gradients, which is formulated as $(x_i - x'_i) \int_0^1 \frac{\partial F(x' + \alpha(x - x'))}{\partial x_i} d\alpha$, where $F(\cdot)$ is a neural network mapping, $x = (x_1, x_2, \ldots, x_N)$ is an input, and $x' = (x'_1, x'_2, \ldots, x'_N)$ is the baseline satisfying $\frac{\partial}{\partial x} F(x)|_{x=x'} = 0$. In practice, the integral can be transformed into a discrete summation $\frac{(x_i - x'_i)}{M} \times \sum_{m=1}^{M} \frac{\partial F(x' + \frac{m}{M}(x - x'))}{\partial x_i}$, where $M$ is the number of steps in the approximation of the integral. S. Singla et al. [160] proposed to use second-order approximations of a Taylor expansion to produce a saliency map so as to consider feature dependencies.

S. Bach et al. [11] proposed layer-wise relevance propagation (LRP) to compute the relevance of one attribute to a prediction by assuming that a model representation $f(x)$ can be expressed as the sum of pixel-wise relevance $R_p^l$, where $x$ is an input image, $l$ is the index of the layer, and $p$ is the index of the pixel of $x$. Thus, $f(x) = \sum_p R_p^L$, where $L$ is the final layer and $R_p^L = \frac{w_p x_p^{L-1}}{\sum_p w_p x_p^{L-1}} f(x)$, where $w_p$ is the weight between pixel $p$ of the $(L-1)^{th}$ layer and the final layer. Given a feed-forward neural network, the pixel-wise relevance score $R_p^1$ of an input is derived by calculating $R_p^l = \sum_j \frac{z_{pj}}{\sum_{p'} z_{p'j}} R_j^{l+1}$ backwards with $z_{pj} = x_p^l w_{pj}^{(l,l+1)}$, where $w_{pj}^{(l,l+1)}$ is the weight between the pixel $p$ of layer $l$ and the pixel $j$ of the $(l+1)^{th}$ layer. Furthermore, L. Arras et al. [9] extended LRP to recurrent neural networks (RNNs) for sentiment analysis. G. Montavon et al. [125] employed the whole first-order term of deep Taylor decomposition to produce a saliency map instead of just gradients. Suppose $\hat{x}$ is a well-chosen root for the function by a model $f(x): f(\hat{x}) = 0$, because $f(x)$ can be decomposed as $f(x) = f(\hat{x}) + \left(\frac{\partial f}{\partial x}|_{x=\hat{x}}\right)^T \cdot (x - \hat{x}) + \epsilon = 0 + \sum_i \frac{\partial f}{\partial x_i}|_{x=\hat{x}} (x_i - \hat{x}_i) + \epsilon$, where $\epsilon$ is high-order terms, the pixel relevance for the pixel $i$ is expressed as $R_i =$

$\frac{\partial f}{\partial x_i}|_{x=\hat{x}}(x_i - \hat{x}_i)$). Inspired by the fact that even though a neuron is not fired, it is still likely to reveal useful information, A. Shrikumar et al. [156] proposed *DeepLIFT* to compute the difference between the activation of each neuron and its reference, where the reference is the activation of that neuron when the network is provided a reference input, and then backpropagate the difference to the image space layer by layer as LRP does. C. Singh et al. [159] introduced contextual decomposition whose layer propagation formula is $\beta_i = W\beta_{i-1} + \frac{|W\beta_{i-1}|}{|W\beta_{i-1}|+|W\gamma_{i-1}|} \cdot b$ and $\gamma_i = W\gamma_{i-1} + \frac{|W\gamma_{i-1}|}{|W\beta_{i-1}|+|W\gamma_{i-1}|} \cdot b$, where $W$ is the weight matrix between the $i^{th}$ and $(i-1)^{th}$ layers and $b$ is the bias vector. The restricting condition is $g_i(x) = \beta_i(x) + \gamma_i(x)$, where $g_i(x)$ is the output of $i^{th}$ layer. $\beta_i(x)$ is considered as the contextual contribution of the input and $\gamma_i(x)$ implies contribution of the input to $g_i(x)$ that is not included in $\beta_i(x)$.

Figure 6 showcases the evaluation of raw gradients, SmoothGrad, IntegratedGrad, and Deep Taylor methods with a LeNet-5-like network. Among them, IntegratedGrad and Deep Taylor methods perform superbly on five digits.

Mutual-information measure to quantify the association between inputs and latent representations of a deep model can also similarly work as the saliency [63], [149], [194]. In addition, there are other methods to obtain saliency maps as well. A. S. Ross et al. [145] defined a new loss term $\sum_i \left(A_i \frac{\partial}{\partial x_i} \sum_{k=1}^{K} \log(\hat{y}_k)\right)^2$ for training, where $i$ is an index of a pixel, $A_i$ is the binary mask to be optimized, $\hat{y}_k$ is the $k^{th}$ digit of the label, and $K$ is the number of class. This loss is to penalize the sharpness of gradients towards a clearer interpretation boundary. R. C. Fong and A. Vedaldi [52] explored to learn the smallest region to delete, which is to find the optimal $m^*$:

$$m^* = \underset{m \in [0,1]^n}{\arg\min} \quad \lambda ||1 - m||_1 + f_c(x_0; m),$$

where $m$ is the soft mask, $f_c(x_0; m)$ represents the loss of the network for an image $x_0$ with the soft mask, and $n$ is the number of pixels. T. Lei et al. [102] utilized a generator to specify segments of an original text as so-called rationales, which fulfill two conditions: 1) rationales should be sufficient as a replacement for the initial text; 2) rationales should be short and coherent. Deriving rationales is actually equivalent to deriving a binary mask, which can be regarded as a saliency map. Based on the above two constraints, the penalty term for a mask is formulated as:

$$\Omega(z) = \lambda_1 ||z||_1 + \lambda_2 \sum_t |z_t - z_{t-1}|,$$

where $z = [z_1, z_2, \dots]$ is a mask, the first term penalizes the number of rationales, and the second term is for smoothness.

The class activation map method (CAM [210]) and its variant [151] utilized global average pooling before a fully connected layer to derive the discriminative area. Specifically, let $f_k(x, y)$ represent the $k^{th}$ feature map, for a given class $c$, the input to the softmax layer is $\sum_k w_k^c \sum_{x,y} f_k(x, y)$, where $w_k^c$ is the weight vector connecting the $k^{th}$ feature map and the class $c$. The discriminative area is obtained as $\sum_k w_k^c f_k(x, y)$, which directly implies the importance of the pixel at $(x, y)$ for class $c$. What's more, some weakly supervised learning methods such as M. Oquab et al. [135] can obtain discriminative areas as well. Specifically, they trained a network only with object labels, however, when they rescaled the feature maps produced by the max-pooling layer, it was surprisingly found that these feature maps were consistent with the locations of objects in the input.

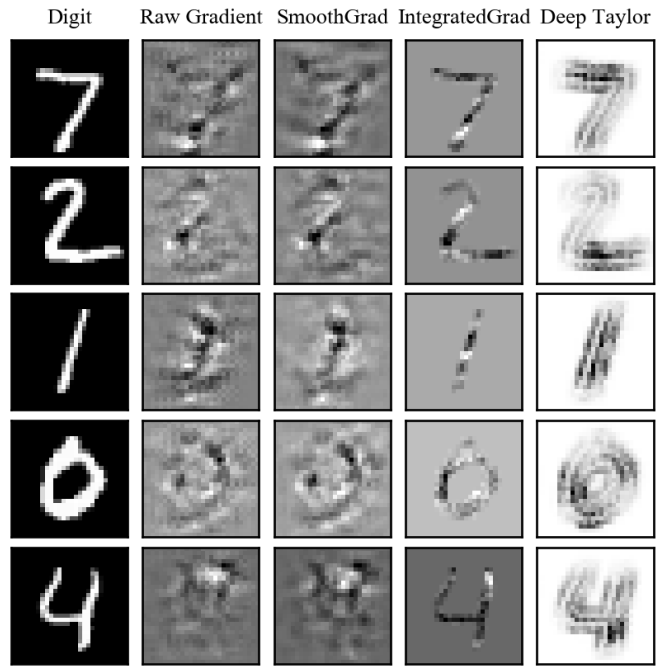

Figure 6. Interpreting a LeNet-5-like network by raw gradient, SmoothGrad, Integrated Gradient, and Deep Taylor methods, respectively. It is seen that Integrated Gradient and Deep Taylor methods have sharper and less noisy saliency map.

· *Proxy*

There are about three ways to prototype a proxy. The first one is direct extraction. The gist of direct extraction is to construct a new interpretable model such as a decision tree [92], [192] or a rule-based system directly from the trained model. As far as the rule extraction is concerned, both decompositional [152] and pedagogical methods [147], [173] can be used. Pedagogical approaches extract rules that enjoy a similar input-output relationship with that of a neural network. These rules do not correspond to the weights and structure of the network. For example, the Validity Interval Analysis (VIA) [118] extracts rules in the following form:

IF (input ∈ a hypercube), THEN class is $C_j$.

R. Setiono and H. Liu [152] clustered hidden unit activation values based on the proximity of activation values. Then, the activation values of each cluster were denoted by their average activation values, at the same time kept the accuracy of the neural network as intact as possible. Next, the input data with the same average hidden unit activation value were clustered together to obtain a complete set of rules. In Figure 7, we illustrate obtained rules from a one-hidden-layer network using R. Setiono and H. Liu's method over the Iris dataset. In a neural network for a binary classification problem, the decision boundaries divide the input space into two parts, corresponding to two classes respectively. The explanation system HYPINV developed in E. W. Saad *et al.* [147] computes for each and every decision boundary hyperplane a tangent vector. The sign of an inner product between an input instance and a tangent vector will imply the position of the input instance relative to the decision boundary. Based on such a fact, a rule system can be established.

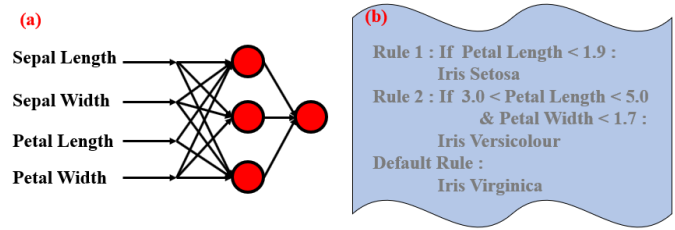

Figure 7. Rule extraction process as proposed by R. Setiono and H. Liu [152]. (a) A one-hidden-layer network with three hidden neurons is constructed to classify the Iris dataset. (b) Rules are extracted via discretizing activation values of hidden units and clustering of inputs, where Petal length and Petal width are dominating attributes for classification of Iris samples. The extracted rules have the same classification performance as that of the original neural network.

Lastly, some specialized networks such as ANFIS [80] and RBF networks [126], straightforwardly correspond to fuzzy logic systems. For example, an RBF network is equivalent to a Takagi-Sugeno rule system [172] that comprises rules such as "if $x \in$ set $A$ and $y \in$ set $B$, then $z = f(x, y)$" [136]. Fuzzy logic interpretation in [48] considers each neuron/filter in a network as a generalized fuzzy logic gate. In this view, a neural network is nothing but a deep fuzzy logic system. Specifically, they analyzed a new type of neural networks, called quadratic networks, in which all the neurons are quadratic neurons that replace the inner product with the quadratic operation [47]. Their interpretation generalized fuzzy logic gates implemented by quadratic neurons, and then computed the entropy based on spectral information of fuzzy operations in a network. It was suggested that such an entropy could have deep connections with properties of minima and the complexity of neural networks.

The second one is called knowledge distillation [23] as Figure 8 shows. Although knowledge distillation techniques are mostly used for model compression, their principles can also be used for interpretability. The motif of knowledge distillation is that cumbersome models can generate relatively accurate predictions, assigning probabilities to all the possible classes, known as soft labels, that are more informative than one-hot labels. For example, a horse is more likely to be classified as a dog instead of a mountain. But with one-hot labeling, both the dog class and mountain class have zero probability. It was shown in [23] that, by the means of matching the logits of the original model, the generalization ability of the original cumbersome model could be transferred into a simpler model. Along this direction, an interpretable proxy model such as a decision tree [38], [186], a decision set [98], a global additive model [171], and a simpler network [75] were developed. For example, S. Tan *et al.* [171] used soft labels to train a global additive model in the form $h_0 +$

$\sum_i h_i(x_i) + \sum_{i \neq j} h_{ij}(x_i, x_j) + \sum_{i \neq j} \sum_{j \neq k} h_{ijk}(x_i, x_j, x_k) + \cdots$ ,
where $\{h_i\}_{i \geq 1}$ could work as a feature saliency directly.

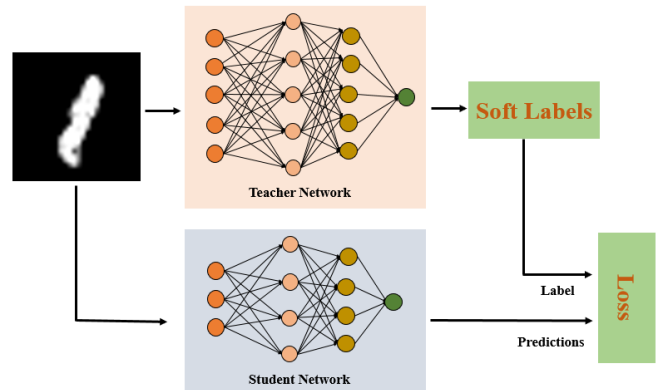

Figure 8. Knowledge distillation is to construct an interpretable proxy by the soft labels from the original complex models.

The last one is to provide a local explainer as a proxy. Local explainer methods locally mimic the predictive behaviors of neural networks. The basic rationale is that when a neural network is inspected globally, it looks complex. However, if we tackle it locally, the picture becomes clearer.

One typical local explainer is Local Interpretable Model-agnostic Explanation (LIME) [141], which synthesizes a number of neighbor instances by randomly setting elements of that sample to zero and computing the corresponding outcomes. Then, a linear regressor is used to fit synthesized instances, where the coefficients of the linear model signify the contributions of features. As Figure 9 shows, the LIME method is applied to a breast cancer classification model to identify which attributes are contributing forces for the model's benign or malignant prediction.

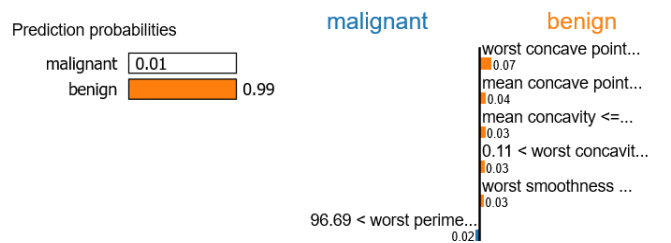

Figure 9. A breast cancer classification task model dissected by LIME. In this case, the sample is classified as benign where worst concave point, mean concave point and so on are contributing forces, while the worst perimeter is the contributing force to drive the model to predict "malignant".

Y. Zhang *et al.* [207] pointed out the lack of robustness in the LIME explanation, which originates from sampling variance, sensitivity to the choice of parameters, and variation across different data points. *Anchor* [142] is an improved extension of LIME, which is to find the most important segments of an input such that the variability of the rest segments does not matter. Mathematically, *Anchor* searches a set: $A = \{z | f(z) = f(x), z \in x\}$, where $f(\cdot)$ is a black-box model, $x$ is the input, and $z$ is the part of $x$. Another proposal LOcal Rule-based Explanation (LORE) was from [64]. The LORE takes advantage of the genetic algorithm to generate the balanced neighbors instead of random neighbors, thereby yielding high-quality training data that alleviates sampling variance of LIME.

· *Advanced Mathematical/Physical Analysis*

Y. Lu *et al.* [114] showed that many residual networks can be explained as discretized numerical solutions of ordinary differential equations, i.e., the inner-working of a residual block in ResNet [69] can be modeled as $u_{n+1} = u_n + f(u_n)$, where $u_n$ is the output of the $n^{th}$ block, and $f(u_n)$ is the block operation. It was noticed that $u_{n+1} = u_n + f(u_n)$ is a one-step finite difference approximation of an ordinary differential equation $\frac{du}{dt} = f(u)$. This idea inspired the invention of ODE-Net [32]. As Figure 10 shows, the starting point and the dynamics are tuned by an ODE-Net to fit a spiral.

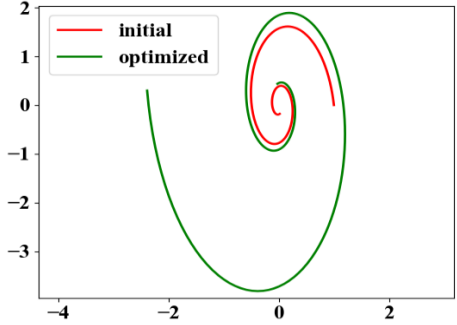

Figure 10. ODE-Net optimizes the start point and the dynamics to fit the spiral shape.

N. Lei *et al.* [101] constructed an elegant connection between the Wasserstein generative adversarial network (WGAN [8]) and optimal transportation theory. They concluded that with low dimensionality hypothesis and the intentionally designed distance function, a generator and a discriminator can exactly represent each other in a closed form. Therefore, the competition between a discriminator and a generator in WGAN in the training is unnecessary.

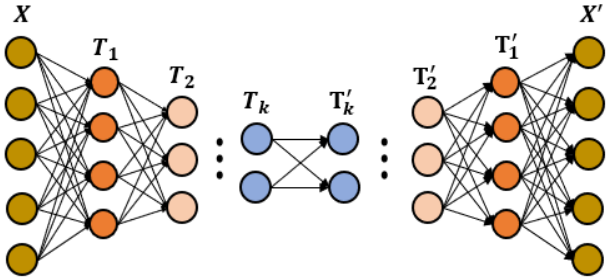

Figure 11: An application of information bottleneck theory to compare mutual information between symmetric layers in an autoencoder.

In [154], it was proposed that the learning of a neural network is to extract the most relevant information in the input random variable $X$ that pertains to an output random variable $Y$. Naively, for a feedforward neural network, the following inequality of mutual information holds:

$$I(Y;X) \geq I(Y;h_j) \geq I(Y;h_i) \geq I(Y;\hat{Y}),$$

where $I(\cdot;\cdot)$ denotes mutual information, $h_i, h_j$ are outputs of hidden layers ($i > j$ means that the $i^{th}$ layer is deeper), and $\hat{Y}$ is a final prediction. Furthermore, S. Yu and J. C. Principe [198] employed an information bottleneck theory to gauge the mutual information states of symmetric layers in a stacked autoencoder as shown in Figure 11:

$$I(X;X') \geq I(T_1;T_1') \geq \cdots \geq I(T_K;T_K').$$

However, it is tricky to estimate the mutual information since the probabilistic distribution of data is usually unknown as a priori.

S. Kolouri *et al.* [91] built an integral geometric explanation for neural networks with a generalized Radon transform. Let $X$ be a random variable for the input, which conforms to the distribution $p_X$, then we can derive a probability distribution function for the output of a neural network $f_\theta(X)$ parametrized with $\theta$: $p_{f_\theta}(z) = \int_X p_X(x)\delta(z - f_\theta(x))\,dx$, which is the generalized Radon transform, and the hypersurface is $H(t,\theta) = \{x \in X | f_\theta(x) = t\}$. In this regard, the transform by a neural network is characterized by the twisted hypersurfaces. H. Huang [77] used the mean-field theory to characterize the mechanism of dimensionality reduction by a deep network that assumes weights in each layer and input data following a Gaussian distribution. In his study, the self-covariance matrix of the output of the $l^{th}$ layer was computed as $C^l$, then the intrinsic dimensionality was defined as $D = \frac{(\sum_{i=1}^{N}\lambda_i)^2}{\sum_{i=1}^{N}\lambda_i^2}$, where $\lambda_i$ is the eigenvalue of $C^l$, and $N$ is the number of eigenvalues. The quantity $D/N$ was investigated across layers to analyze how compact representation are learned across layers. J. C. Ye *et al.* [193] utilized a framelet theory and low-rank Hankel matrix to represent signals in terms of their local and non-local bases, corresponding to convolution and generalized pooling operations. However, in their study the network structure was simplified in concatenating two ReLU units into a linear unit such that the nonlinearity from ReLU units could be circumvented. As far as advanced physic models are concerned, P. Mehta and D. C. Schwab [121] built an exact mapping from the Kadanoff variational renormalized group [82] to the restricted Boltzmann Machine (RBM) [148]. This mapping is independent of forms of the energy functions and can be scaled to any RBM.

Theoretical neural network studies are essential to interpretability as well. Currently, theoretical foundations of deep learning are primarily from three perspectives: representation, optimization, and generalization.

*Representation*: Let us include two examples here. The first example is to explain why deep networks is superior to shallow ones. Recognizing success of deep networks, L. Szymanski and B. McCane [170], D. Rolnick and M. Tegmark [144], N. Cohen *et al.* [37], H. N. Mhaskar and T. Poggio [124], R. Eldan and O. Shamir [44], and S. Liang and R. Srikant [109] justified that a deep network is more expressive than a shallow one. The basic idea is to construct a special class of functions that can be efficiently represented by a deep network but hard to be approximated by a shallow one. The second example is to understand utilities of shortcut connections of deep networks. A. Veit *et al.* [178] showed that residual connections can render a neural network to manifest an ensemble-like behavior. Along this direction, it was reported in [110] that with shortcuts, a network can be super slim to allow for universal approximation.

*Optimization*: Generally, optimizing a deep network is a NP-hard non-convex problem. The pervasive existence of saddle points [56] leads to that even finding a local minimum is also NP-hard [5]. Of particular interest to us is why an over-parametrized network can still be optimized well because a deep network is a kind of over-parametrized networks. The character of an over-parameterized network is that the number of parameters in a network exceeds the number of data instances. M. Soltanolkotabi *et al.* [163] showed that when data are Gaussian distributed and activation functions of neurons are quadratic, the landscape of an over-parameterized one-hidden-layer network allows global optimum to be searched efficiently. Q. Nguyen and M. Hein [130] demonstrated that with respect to linearly separable data, under assumptions on the rank of weight matrices of a feedforward neural network, every critical point of a loss function is a global minimum. Furthermore, A. Jacot *et al.* [78] showed that when the number of neurons in each layer of a neural network goes infinitely large, the training only renders small changes for the network function. As a result, the training of the network turns into the kernel ridge regression.

*Generalization*: Conventional generalization theory is incompetent to explain why a deep network can generalize well despite that the number of parameters of a deep network is many more than the number of samples. Recently proposed generalization bounds [127] that rely on the norm of weight matrices partially solved this problem. However, these bounds have an abnormal dependence on data that more data lead to a larger generalization bound, which apparently contradicts the common sense. We prospect that more efforts are needed to resolve the generalization puzzle satisfactorily [18], [122].

· *Explaining-by-Case*

Basically, case-based explanations present a case that is believed by a neural network to be most similar to the query case needing an explanation. Finding a similar case for explanation and selecting a representative case from data as the prototype [19] are basically the same thing and just use different metrics for similarity. While prototype selection is to find a minimal subset of instances that can represent the whole dataset, case-based explanations use the similarity metric based on the closeness of representations of a neural network, thereby exposing the hidden representation information of the neural network. In this light, case-based explanations are also related to deep metric learning [150].

As shown in Figure 12, E. Wallace *et al.* [181] employed the k-nearest neighbor algorithm to obtain the most similar cases for the query case in the feature space and then computed the percentage of the nearest neighbors belonging to the expected class as a measure for interpretability, suggesting how much a prediction is supported by data. C. Chen *et al.* [31] constructed a model that could dissect images by finding prototypical parts. Specifically, the pipeline of the model splits into multiple channels after convolutional layers, in which the function of each channel is expected to learn a prototypical part of the input such as the head or body of a bird. The decision for an input image is made based on the similarity of features of channels.

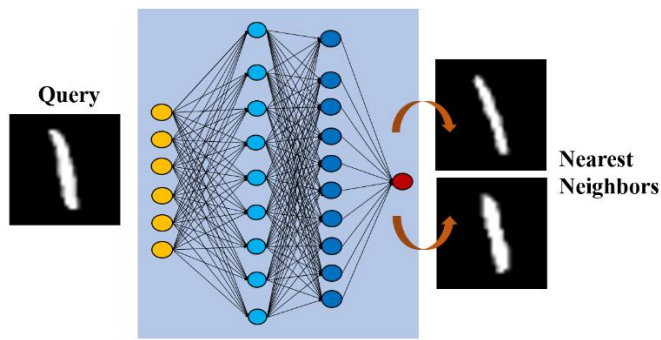

Figure 12. Explaining-by-case presents the nearest neighbors in response to a query.

S. Wachter *et al.* [180] offered a novel case-based explanation method by providing a counterfactual case, which is an imaginary case that is close to the query but has a different output from that of the query. Counterfactual explanation provides the so-called "closest possible case" or the smallest change to yield a different outcome. For example, counterfactual explanations may produce the following statement: "If you have a good striker, your team would have won this soccer game." Coincidently, techniques to generate a counterfactual explanation have been developed for the purpose of "adversarial perturbation", i.e., structural attack [191]. Essentially, finding a closest possible case $x'$ to the input $x$ is equivalent to finding the smallest perturbation to $x$ such that the classification result changes. For example, the following optimization can be built:

$$\arg\min_{x'} \ \lambda(f(x') - y')^2 + d(x, x'),$$

where $\lambda$ is a constant, $y'$ is a different label, and $d(\cdot, \cdot)$ is chosen to be the Manhattan distance in hope that the input be minimally perturbed. Y. Goyal *et al.* [62] explored an alternative way to derive a counterfactual visual explanation. Given an image $I$ with a label $c$, since the counterfactual visual explanation represents the change for the input that can force the model to yield a different prediction class $c'$, they selected an image $I'$ with a label $c'$ and managed to recognize the spatial region in $I$ and $I'$ such that the replacement of the recognized region would alter the model prediction from $c$ to $c'$.

· *Explaining-by-Text*

Neural image captioning uses a neural network to produce a natural language description for an image. Despite that neural image captioning is initially not for network interpretability, descriptive language about images can tell the information about how a neural network analyzes an image. One representative method is from [84] that combines a convolutional neural network and a bidirectional recurrent neural network to obtain a bimodal embedding. Due to the hypothesis that the two embeddings representing similar semantics across two modalities should share the nearby locations of two spaces, the objective function is defined as

$$S_{IT} = \sum_{t \in g_T} \max_{i \in g_I} v_i^T s_t,$$

where $v_i$ is the $i^{th}$ image fragment in the set $g_I$, and $s_t$ is the $t^{th}$ word in a sentence $g_T$. Another representative method is the attention mechanism [137], [179], [189], [190], where deep features are to align the corresponding text descriptions by a recursive neural network such as LSTM [73]. An explanation for deep features is provided by the corresponding words in the text and attention maps, which reflect which parts of an image attract the attention of the neural network.

As shown in Figure 13, in the $k^{th}$ attention module that takes $y_0, y_1, \ldots, y_n$ as input, suppose its output is $t_k = \sum_i y_i s_{ki}$. $s_{k0}, s_{k1}, \ldots, s_{kn}$ together form an attention map for $t_k$ with respect to the associated word. However, S. Jain and B. C. Wallace [79] argued that an attention map is not qualified to work as an explanation because they observed that the attention map was not correlated with other importance measures of features such as gradient-based measures, and the change of attention weights yielded no changes in prediction.

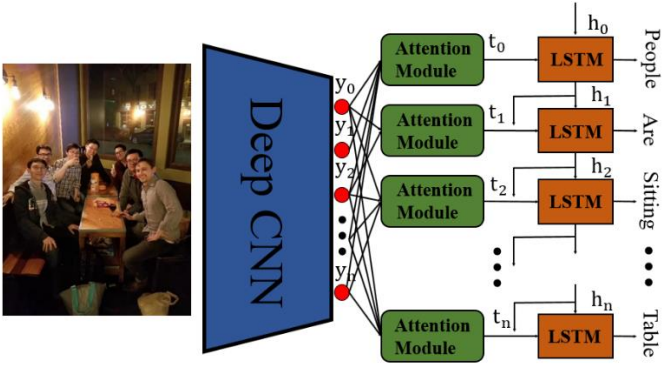

Figure 13. Image captioning with attention modules provides an explanation to the features mined by a deep convolutional network.

### C. Ad-hoc Interpretable Modeling

· *Interpretable Representation*

Traditionally, regularization techniques for deep learning are primarily designed to avoid overfitting. However, it is also feasible to devise regularization techniques to enhance an interpretable representation in terms of decomposability [33], [165], [182], [205], monotonicity [195], non-negativity [34], sparsity [167], human-in-the-loop prior [96], and so on.

For example, X. Chen *et al.* [33] invented *Info*GAN which is a simple but effective way to learn an interpretable representation. Traditionally, a generative adversarial network (GAN) [60] imposes no restrictions on how a generator utilizes the noise. In contrast, *Info*GAN maximizes the mutual information between the latent codes and observations, forcing each dimension of noise to encode a semantic concept. Particularly, the latent codes are made of discrete categorical codes and continuous style codes. As shown in Figure 14, two style codes control the localized part and the digit rotation respectively.

Incorporating monotonicity constraints [195] is also useful to enhance interpretability. A monotonical relationship means when the value of a specified attribute increases, the predictive value of a model either increases or decreases. Such a simplicity promotes interpretability as well. J. Chorowski and J. M. Zurada [34] imposed non-negativity to weights of neural networks and argued that it could improve interpretability because it eliminated the cancellation and aliasing effects among neurons. A. Subramanian *et al.* [167] employed a k-sparse autoencoder for word embedding to promote sparsity in the embedding and claimed that this enhanced interpretability because a sparse embedding reduced the overlap between words. I. Lage *et al.* [96] proposed a novel human-in-the-loop evaluation in selecting a model. Specifically, a diverse set of models were trained and sent to users for evaluation. Users were asked to predict what the label of a data point would be assigned by a model $M$. The shorter the response time was, the better a user understood the model. Then, the model with the lowest response time was chosen.

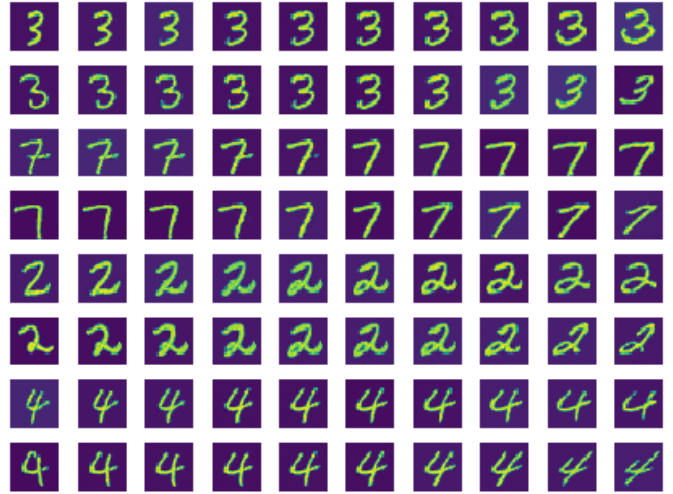

Figure 14. In an *Info*GAN, two latent codes control the localized parts and rotation parts respectively.

· *Model Renovation*

L. Chu *et al.* [35] proposed to use piecewise linear functions as activations for a neural network (PLNN), thereby the decision boundaries of PLNN could be explicitly defined and further a closed-form solution could be derived for predictions of a network. As Figure 15 shown, F. Fan *et al.* [49] proposed Soft-Autoencoder (Soft-AE) by using adaptable soft-thresholding units in encoding layers and linear units in decoding layers. Consequently, Soft-AE can be interpreted as a learned cascaded wavelet adaptation system.

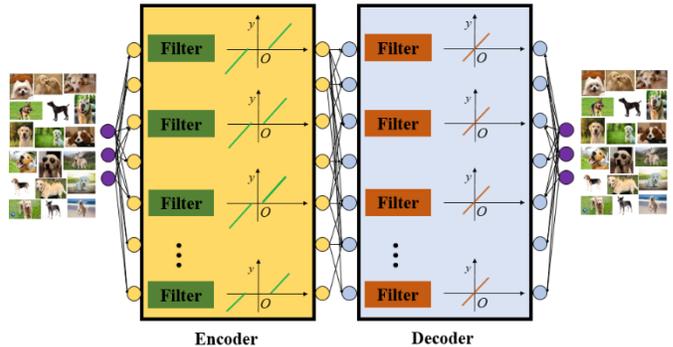

Figure 15. Soft-autoencoder with soft-thresholding functions as activation functions in the encoding layers and the linear function as activations in the decoding layers, thereby admitting a direct correspondence to the wavelet adaptation system.

L. Fan [50] explained a neural network as a generalized Hamming network, whose neurons compute the generalized Hamming distance: $h(\boldsymbol{x}, \boldsymbol{w}) = \sum_{l=1}^{L} w_l + \sum_{l=1}^{L} x_l - 2\boldsymbol{x} \cdot \boldsymbol{w}$ for an input $\boldsymbol{x} = (x_1, \ldots, x_L)$ and a weight vector $\boldsymbol{w} = (w_1, \ldots, w_L)$. The bias term in each neuron is specified as $b = -\frac{1}{2}(\sum_{l=1}^{L} w_l + \sum_{l=1}^{L} x_l)$ so that each neuron is a generalized Hamming neuron. In this regard, the function of the batch normalization is demystified as making the bias suitable for computation of the generalized Hamming distance. C. C. J. Kuo *et al.* [95] proposed a transparent design for constructing a feedforward convolutional network without the need of backpropagation. Specifically, filters in convolutional layers were built by selecting principal components of PCA for outputs of earlier pooling layers. A fully connected layer was constructed by treating it as a linear-squared regressor.

D. A. Melis and T. Jaakkola [123] claimed that a neural network model $f$ is interpretable if it has the form that $f(x) = g(\theta_1(x)h_1(x), \ldots, \theta_k(x)h_k(x))$, where $h_i(x)$ is the prototypical concept from the input $x$ and $\theta_i(x)$ is the relevance associated with that concept, $g$ is monotonic and completely additively separable. Such a model can learn interpretable basis concepts and facilitate saliency analysis. Similarly, J. Vaughan *et al.* [177] designed a network structure to compatibly learn the function formulated as $f(x) = \mu + \gamma_1 h_1(\beta_1^T x) + \gamma_2 h_2(\beta_2^T x) + \cdots + \gamma_K h_K(\beta_K^T x)$, where $\beta_k$ is the projection, $h_k(\cdot)$ represents the nonlinear transformation, $\mu$ is the bias, and $\gamma_k$ is the weighting factor. Such a model is more interpretable than a general network, because the function of this model has simpler partial derivatives that can simplify saliency analysis, statistical analysis, and so on.

C. Li *et al.* [104] proposed deep supervision by using prior hierarchical tasks on features of intermediate layers. Specifically, we have a dataset $\{(x, y_1, \ldots, y_m)\}$, where labels $y_1, \ldots, y_m$ are hierarchical that $y_j, j < i$ is a strict necessary condition for the existence of $y_i, i > 1$. Such a scheme introduces a modularized idea that through supervision of a specific task for an intermediate layer, the learning of that layer is steered towards the pre-specified task, thereby gaining interpretability.

T. Wang [183] proposed to use an interpretable and insertable substitute on a subset of data which the complex black-box model overkills. In their work, a rule set was built as an interpretable model to make a decision on the input data first. Those inputs which a rule set was handicapped to classify were passed into the black-box model for decision making. The logic of this hybrid predictive system is that an interpretable model for regular cases without compromising accuracy, a complex black-box model for complicated cases.

C. Jiang *et al.* [81] proposed finite automata-recurrent neural network (FA-RNN) that can be directly transformed into the regular expressions such that a good interpretability is extracted. The roadmap is that the constructed FA-RNN can be approximated into finite automata, and further transformed into regular expressions because finite automata and a regular expression are mutually convertible. In analogy, a regular expression can also be decoded into an FA-RNN as an initialization. FA-RNN is a good example to manifest the synergy between a rule system and a neural network.

## III. INTERPRETABILITY IN MEDICINE

These days, reports are often seen in the news that deep learning-based algorithms outperform experts or classic algorithms in the field of medicine [153]. Indeed, given an adequate computational power and well-curated datasets, a properly designed model can deliver competitive performance in most well-defined pattern recognition tasks. However, due to the high stakes of medicine-concerned applications, it is not sufficient to have a deep learning model that produces correct answers without an explanation. In this section, we focus on several exemplary papers concerning applications of interpretability methods in medicine, and we organize the articles of relevance in accordance with the aforementioned taxonomy.

· *Post-hoc Interpretability Analysis*

*Feature analysis*

P. Van Molle *et al.* [176] visualized convolutional neural networks to assist decision-making for skin lesion classification. In their work, feature activations generated from the last two convolutional layers were rescaled to the size of an input image as the activation maps. Where a map has high activations were inspected. The activation strengths across different border types, skin colors, skin types, etc. were compared. The activation map exposed a risk that some unexpected regions had uncommonly high activations.

D. Bychkov *et al*. [24] utilized a model that combines a VGG-16 network [158] and an LSTM network [73] to predict five-year survival of colorectal cancer based on digitized tumor tissue samples. In their work, an RGB pathological image was split into many tiles. A VGG-16 network extracted a high-dimensional feature vector from each tile, which was then fed into an LSTM network to predict five-year survival. They used t-SNE [116] to map features learned by VGG-16 into a two-dimensional space for visualization and found that different classes of features of VGG-16 were well separated.

*Saliency*

I. Sturm *et al*. [166] applied a deep network with LRP [11] for the single-trial EEG [22] classification. The network entails two linear mean pooling layers before being activated or normalized. The feature importance score is assigned by LRP (S. Bach *et al.*, 2015).

J. R. Zech *et al*. [200] developed a deep learning model for chest radiography to classify patients into having pneumonia or not. Through interpretability analysis by CAM [210], they reported the risk that a deep learning model could make an incorrect decision by capturing features irrelevant to diseases, such as metal tokens.

O. Oktay *et al*. [134] combined attention gates with the decoder part of U-Net to cope with interpatient variation in organs' shapes and sizes. The proposed model can improve model sensitivity and accuracy by inhibiting representations of irrelevant regions. Aided by attention gates, they found that the model gradually shifted its attention to regions of interest.

D. Ardila *et al.* [7] proposed a deep learning algorithm that considers a patient's current and previous CT volumes to predict the risk of lung cancer. They used the integrated gradient method [168] to derive saliency maps and invited experienced radiologists to examine the fidelity of these maps. It turned out that in all cases, the readers strongly agreed that the model indeed focused on the nodules.

H. Lee *et al.* [100] reported an attention-assisted deep learning system for detection and classification of acute intracranial haemorrhage, where an attention map identified a region relevant to the disease. They evaluated the localization accuracy of the attention maps by computing the proportion of bleeding points overlapping with the attention maps. Overall, it was found that 78.1% bleeding points were detected in the attention maps.

W. Caicedo-Torres and J. Gutierrez [25] proposed a multi-scale deep convolutional neural network for the mortality prediction based on the measurement of 22 different items in ICU such as the sodium index, urine output, etc. In their work, three temporal scales were represented by stacking convolutional kernels of dimensions $3 \times 1$, $6 \times 1$, and $12 \times 1$. The saliency map by DeepLIFT [156] was utilized for interpretability.

H. Guo *et al.* [66] introduced an effective dual-stream network that conjugates extracted features from ResNet [69] and clinical prior knowledge to predict the mortality risk of patients based on low-dose CT images. To further testify the effectiveness of the proposed model, they utilized t-SNE [116] to reduce the dimensionality of feature maps of malignant and benign samples and found that malignant and benign features were well separated. Also, they applied CAM [210] to reveal that the deceased subjects correctly classified by the model were prone to have strong activations.

*Proxy*

Z. Che *et al.* (2016) applied knowledge distillation into a deep model to learn a gradient boosting tree [106] (GBT), that provides not only robust prediction performance but also a good interpretability in the context of electronic health record prediction. Specifically, they trained three deep models respectively, and then used predictions of deep models as labels to train a GBT model. Experiments on a Pediatric ICU dataset were reported that the GBT model maintained the prediction performance of deep models in terms of mortality and ventilator-free days.

S. Pereira *et al.* [138] combined global and local interpretation efforts for brain tumor segmentation and penumbra estimation in stroke lesions, where the global interpretability was derived from mutual information to sense the dependence between an input sample and the prediction, while the local interpretability was cast by a variant of LIME [141].

*Explaining-by-Case*

N. C. F. Codella *et al.* [36] employed saliency and explaining-by-case methods to explain a dermoscopic image analysis network which was jointly trained by disease labels with a triple-let loss. Specifically, the interpretability was gained by the discovered neighbors and localized regions that were most relevant to the distance from queries and neighbors.

*Explaining-by-Text*

Z. Zhang *et al.* [208] proposed an all-in-one network that read pathology bladder cancer images, generated diagnostic reports, retrieved images according to symptomatic descriptions, and visualized attention maps. They designed an auxiliary attention sharpening module to improve the discriminability of attention maps. Pathologists' feedbacks suggested that the explanatory maps tended to highlight regions that concern with carcinoma-informative regions.

· *Ad-hoc Interpretable Modeling*

*Interpretable Representation*

X. Fang and P. Yan [51] devised the Pyramid Input Pyramid Output Feature Abstraction Network (PIPO-FAN) with multiple arms for multi-organ segmentation. Each of the arm handles the information on one scale. The total loss is obtained by adding the segmentation loss to each of these arms such that segmentation-wise features are generated in each arm. Visualization analysis suggested that features from different arms have hierarchical semantical meanings, i.e., some are blurry but contain global class-wise information, while the others contain local boundary information. As shown in Figure 16, the segmentation loss creates semantically meaningful features, where low-scale arms produce more details and high-scale arms find global morphologies.

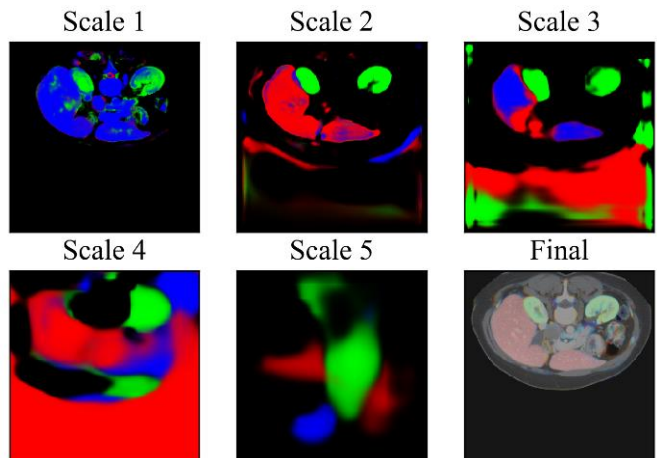

Figure 16. Visualization of feature maps of different arms in PIPO-FAN, where low-scale sub-networks produce local structural details and high-scale sub-networks target global morphological information.

*Model Renovation*

W. Gale *et al.* [55] combined a DenseNet (G. Huang *et al.*, 2017) model with an LSTM model [73] for detection of hip features from pelvic X-ray radiographs. A radiologist hand-labelled standard descriptive terms to construct a semantic dataset for these radiographs. Their model consistently generated informative sentences favored by doctors over saliency maps. Also, they demonstrated that the combination of visualization and text interpretation give an interpretation superior to either of them alone.

C. Biffi *et al.* [20] employed a variational autoencoder [87] (VAE)-based model for classification of cardiac diseases as well as structurally remodeling based on cardiovascular images. In their scheme, registered left ventricular (LV) segmentations at end-diastolic (ED) and end-systolic (ES) phases were encoded in a low-dimensional latent space by VAE. The learned latent low dimensional manifold was connected to a multilayer perceptron (MLP) for disease classification. The interpretation was given by an activation maximization technique. The "deep dream" of MLP was derived and inverted to the image space for visualization.

S. Shen *et al.* [155] built an interpretable deep hierarchical semantic convolutional neural network (HSCNN) to predict the malignancy of pulmonary nodules in CT images. HSCNN consists of three modules: a general feature learning module, a low-level task module that predicts semantic characteristics such as sphericity, margin, subtlety, and so on, and a high-level task module absorbs information from both general features and low-level task predictions to produce an overall lung nodule malignancy. Due to the semantic meaning contained in the low-level task, HSCNN has boosted interpretability.

Z. Zhang *et al.* [209] developed a deep convolutional network to automate the whole-slide reading of pathology images for tumors and the diagnosis process of pathologists. Specially, the network can generate a clinical pathology report along with attention-assisted features.

Y. Lei *et al.* [103] observed that CAM [210] and Grad-CAM [151] are for interpreting localization tasks and tend to ignore fine-grained structures. Consequently, they proposed a shape-and-margin-aware soft activation map (SAM) that could probe subtle but critical features in a lung nodule classification task. The comprehensive experimental comparisons showed that compared to CAM and Grad-CAM, SAM can reveal relatively discrete and irregular features around nodules.

## IV. PERSPECTIVE

In this section, we suggest a few directions, in hope to advance the understanding and practice of artificial neural networks.

· *Synergy of Fuzzy Logic and Deep Learning*

Fuzzy logic [199] was a buzz phrase in the last nighties. It extends the Boolean logic from 0-1 judgement to imprecise inference with fuzziness in the interval [0, 1]. Fuzzy theory can be divided into two branches: fuzzy set theory and fuzzy logic theory. The latter, with an emphasis on "IF-THEN" rules, has demonstrated effectiveness in dealing with a plethora of complicated system modeling and control problems. Nevertheless, a fuzzy rule-based system is restricted by the acquisition of a large number of fuzzy rules, a process that is tedious and computationally expensive. While a neural network is a data-driven method that extracts knowledge from data through training, with the knowledge represented by neurons in a distributed manner. However, a neural network falls short of delivering a satisfactory result in the context of small data and suffers from the lack of interpretability. In contrast, a fuzzy logic system employs experts' knowledge and represents a system in the form of IF-THEN rules. Although a fuzzy logic system merits interpretability and accountability, it is incompetent in efficient and effective knowledge acquisition. It seems that a neural network and a fuzzy logic system are complementary to each other. Therefore, it is instrumental to combine the best of two worlds towards an enhanced interpretability. In fact, this roadmap is not totally new. There have been several combinations along this direction: ANFIS model [80], generic fuzzy perceptron [126], RBF networks [21], and so on.

One suggestion is to build a deep RBF network. Given the input vector $\boldsymbol{x} = [x_1, x_2, ..., x_n]$, an RBF network is expressed as $f(\boldsymbol{x}) = \sum_i^n w_i \phi_i(\boldsymbol{x} - \boldsymbol{c}_i)$, where $\phi_i(\boldsymbol{x} - \boldsymbol{c}_i)$ is usually selected as $\exp\left(-\frac{||\boldsymbol{x}-\boldsymbol{c}_i||^{\wedge}2}{2\sigma^2}\right)$, where $\boldsymbol{c}_i$ is the cluster center of the $i^{th}$ neuron. It was proved the functional equivalence between an RBF network and a fuzzy inference system under mild conditions [21]. Also, an RBF network is shown to be a universal approximator [136]. Hence, an RBF network is a potentially sound vehicle that can encode fuzzy rules into its adaptive representation without loss of accuracy. Reciprocally, rule generation and fuzzy rule representation in an adaptable RBF network are more straightforward compared to a multilayer perceptron. Although current RBF networks are of one-hidden-layer structures, it is feasible to develop deep RBF networks, which can be viewed as a deep fuzzy rule system. A greedy layer-wise training algorithm was developed in [71], which successfully solved the training problem for deep networks. It is possible to translate such success into the training of deep RBF networks. Then, the correspondence between a deep RBF network and a deep fuzzy logic system will be applied to obtain a deep fuzzy rule system. We believe that efforts should be made to synergize fuzzy logic and deep learning techniques aided by big data along this direction.

· *Convergence of Neuroscience and Deep Learning*

Up to date, truly intelligent systems are still only human. The artificial neural networks in their earlier forms were clearly inspired by biological neural networks [120]. However subsequent developments of neural networks were, to a much less degree, pushed by neurological and biological insights. As far as interpretability is concerned, since biological and artificial neural networks are deeply connected, advances in neuroscience should be relevant and even instrumental to the development and interpretation of deep learning techniques. We believe that the neuroscience promises a bright future of deep learning interpretability in the following aspects.

*Cost function.* The effective use of cost functions is a key driving force for the development of deep networks in the past years; for example, the adversarial loss used in GANs [60]. In previous sections, we have highlighted cases which demonstrate that an appropriate cost function will enable a model to learn an interpretable representation, such as enhance feature disentanglement. Along this direction, a myriad of cost functions can be built to reflect biologically plausible rationales. Indeed, our brain can be modeled as an optimization machine [119], which has a powerful credit assignment mechanism to form a cost function.

*Optimization algorithm.* Despite the huge success achieved by backpropagation, it is far from ideal in the view of neuroscience. Truly in many senses, backpropagation fails to manifest the true behaviors of how a human neural system tunes the synapses of a neuron. For example, in a biological neural system, synapses are updated in a local manner [94] and only depend on the activities of presynaptic and postsynaptic neurons. However, connections in deep networks are tuned through non-local backpropagation. Figure 17 shows a bio-plausible learning algorithm for a two-layer network on CIFAR-100 [93]. Additionally, a neuromodulator is missing in deep networks in contrast to the inner-working of a human brain, where the state of one neuron can exhibit different input-output patterns controlled by a global neuromodulator like dopamine, serotonin, and so on [162]. Neuromodulators are believed to be critical due to their ability to selectively control on and off states of one neuron which is equivalently switching the involved cost function [13].

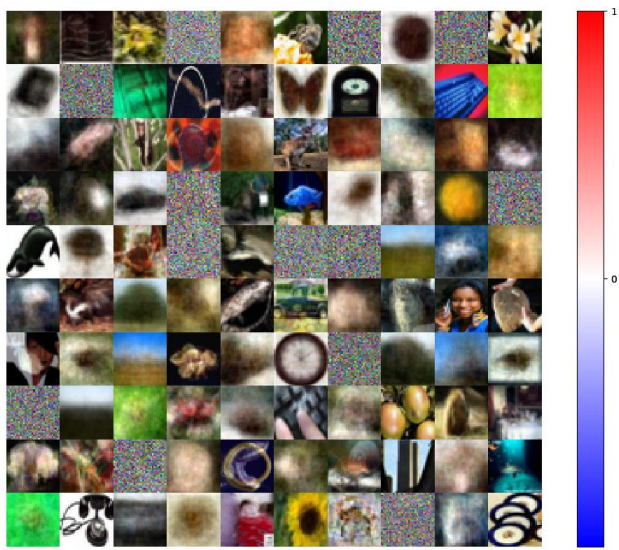

Figure 17. Visualization of weights of a network learned by a bio-plausible algorithm, where prototypes of training image are captured [94].

Considering that there are quite few studies discussing the interpretability of training algorithms, powerful and interpretable training algorithms will be highly desirable. Just like for classic optimization methods, we wish that future non-convex optimization algorithms will have some kinds of uniqueness, stability, and continuous dependency on data, etc.

*Bio-Plausible Architectural Design.* In the past decades, neural networks were designed in diverse architectures from simple feedforward networks to deep convolutional networks and other highly sophisticated networks. The structure determines functionality, i.e., a specific network architecture regulates the information flow with distinct characteristics. Therefore, specialized architectures are useful as effective solutions for intended problems. Currently, the structural differences between deep learning and biological systems are eminent. A typical network is used and tuned for most tasks based on big data, while a biological system learns from a small number of data and generalizes very well. Clearly, a huge amount of knowledge needs to be learned from biological neural networks so that a more desirable and explainable neural network architectures can be designed.

· *Interpretability in Medicine*

A majority of interpretability research efforts in medicine are only for classification tasks, but radiological practices cover a large variety of tasks such as image segmentation, registration, reconstruction, and so on. Clearly, interpretability is also closely relevant to these areas, and therefore it is in need to promote interpretability research in these domains. On the one hand, more efforts should be made to extend the existing interpretation methods to other tasks that have not been explored. On the other hand, practitioners can design task-specific interpretation methods with their expertise and insights. For example, in image segmentation, explaining why a voxel receives a class label in image segmentation is much harder than explaining which area in the input image is responsible for a prediction in image classification. Similarly, for image reconstruction, interpretability could be quite complicated. In this regard, our recently proposed ACID framework allows a synergistic integration of data-driven priors and compressed sensing (CS)-modeled priors, enforcing both of which iteratively via physics-based analytic mapping [188]. By doing so, modern CS and state-of-the-art deep networks are united to overcome the vulnerabilities of existing deep reconstruction networks, at the same time transferring the interpretability of the model-based methods to the hybrid deep neural networks.

In addition to the above referenced publications, gaining interpretability ultimately also relies on medical doctors, who have invaluable professional training despite some biases and errors. As a result, active collaboration among medical doctors, technical experts, and theoretical researchers to design effective, efficient, and reproducible ways to assess and apply interpretability methods will be an important avenue for future development of deep learning methods.

## V. CONCLUSION

In conclusion, we have reviewed key ideas, implications, limitations of existing interpretability studies, and illustrated some typical interpretation methods through examples. In doing so, we have depicted a holistic landscape of interpretability research using the proposed taxonomy and introduced applications of interpretability in medicine particularly. Figures 3, 5, 6, 7, 9, 10, 16, 17 are visualization results from our own implementation of chosen interpretation methods. We have open-sourced relevant codes in the GitHub (https://github.com/FengleiFan/IndependentEvaluation). There is no doubt that a unified and accountable interpretation framework is critical to elevate interpretability research into a new phase. In the future, more efforts are needed to reveal the essence of deep learning. Because this field is still highly interdisciplinary and rapidly evolving, there are great opportunities ahead that will be both academically and practically rewarding.

## VI. ACKNOWLEDGEMENT

The authors are grateful for Dr. Hongming Shan's suggestions (Fudan University) and anonymous reviewers' advice.